# Clinical Harness for Governable Medical AI Skill Ecosystems

Tianhan Xu[1], Lei Bao[1], Zhe Hu[2], Tian Shen[3,*], Yongxiang Wang[4,*]

1 Department of Medical Informatics, Northern Jiangsu People's Hospital, Clinical Teaching Hospital of Medical School, Nanjing University, Yangzhou, China.

2 Department of Computing, The Hong Kong Polytechnic University, Hong Kong Special Administrative Region of China.

3 SenseTime Research, Shanghai, China.

4 Department of Orthopedics, Northern Jiangsu People's Hospital, Clinical Teaching Hospital of Medical School, Nanjing University, Yangzhou, China.

First author email: xth925044@yzu.edu.cn

*Correspondence: Tian Shen, shentian@sensecare.com; Yongxiang Wang, wangyongxiang@nju.edu.cn

**Abstract:** Medical AI remains organized around isolated models, whereas care requires accountable capabilities that persist across time. We define clinical AI skills and propose the Clinical Harness, a runtime governance architecture that registers, orchestrates, constrains and monitors them. Using osteoporosis as an exemplar, we show how knowledge-driven, data-driven and physics-enhanced skills can support lifecycle care and provide a governed substrate for future medical agents.

## 1. Introduction: medical AI's translational gap is also an organizational gap

AI has advanced rapidly across health and medicine [1,2]. Deep-learning systems can support imaging interpretation and opportunistic screening [3–5], while large language models (LLMs) and retrieval-augmented generation (RAG) are being explored for patient education, guideline interpretation and decision support [6–8]. Yet regulatory approval or strong task-level performance does not guarantee clinical benefit. External validation, workflow integration and post-deployment monitoring remain uneven, and evidence of improved patient outcomes is limited [9–11].

This translational gap is also organizational. Most medical AI is built as a task-specific model: trained on one modality, optimized for one endpoint and evaluated with model-level metrics. Care is organized differently. It is longitudinal and collaborative; evidence, uncertainty, responsibility and patient preferences move across visits, disciplines and interventions. A clinically useful capability must therefore persist beyond a single inference and remain interpretable when handed from one person or system to another.

The organizational problem appears after deployment as much as before it. Different models may use incompatible units, populations, thresholds or versions; an apparently reasonable sequence of individually accurate outputs can therefore produce an unsafe pathway. Conventional validation usually treats these components independently, whereas clinical risk often emerges at their interfaces. The missing object is not another general model, but an explicit unit of capability and a runtime mechanism for governing how such units interact.

Osteoporosis makes this mismatch visible. Its pathway spans prevention, opportunistic detection, diagnostic confirmation, fracture-risk assessment, pharmacological treatment, vertebral augmentation, rehabilitation and surveillance. A model may estimate bone mineral density or detect a vertebral fracture, but it cannot by itself decide when it should run, whether its input is valid, how its output should change care, or what should happen when it fails.

Medical AI deployment is therefore a problem of capability organization. We define a clinical AI skill as the clinically meaningful unit between an isolated model and a medical agent, and the Clinical Harness as the runtime infrastructure that registers, orchestrates, constrains and monitors such skills. Osteoporosis illustrates how knowledge-driven, data-driven and physics-enhanced skills can be coordinated across lifecycle care. A skill defines the permitted clinical capability; the Clinical Harness governs its execution, review, audit and updating.

## 2. Clinical AI skills: an abstraction between models and agents

A unit of AI capability must carry clinical meaning. The conventional unit, the model, is semantically incomplete: a fracture-risk model may output a probability, but does not define its intended-use population, contraindications, evidence source, fallback strategy, human-review requirement or monitoring status.

We define a clinical AI skill as a software-defined clinical capability that performs a specified task in a defined context and carries explicit evidence, boundaries, safety constraints, human oversight and governance metadata. A skill may contain one or more models, rules, knowledge bases, simulation modules, interface contracts and escalation logic. It is not synonymous with an application programming interface, plug-in or model endpoint: those describe technical access, whereas a clinical AI skill defines the conditions under which an output may enter care.

Six properties distinguish a skill from a model. It is goal-directed, mapping to a clinical decision node; boundary-aware, declaring intended use and out-of-scope inputs; evidence-aware, linking outputs to guidelines, validation studies or simulation evidence; safety-constrained, withholding or escalating when inputs are invalid or confidence is low; collaborative, preserving clinician authority; and governable, with version control, monitoring, audit trails and revalidation.

A clinical AI skill card standardizes registration. It records the interface and operating boundary [12–14], evidence and validation status [15–17], and known failure modes, bias risks and review requirements [18]. The card serves as both a technical contract and a clinical instruction label. MLOps can manage model versions and deployment, and a workflow engine can route tasks; the skill card links these functions to clinical intent, evidence and accountability. Supplementary Table 1 specifies, for S1-S9, the clinical objective, intended-use population, inputs, outputs, operating boundary, exclusions or contraindications, validation evidence, safety gates and fallback, human oversight, and audit and monitoring requirements.

Skill composition is conditional rather than automatic. Two individually validated skills may be incompatible if they assume different populations, time windows, anatomical definitions or outcome labels. A composed pathway therefore needs compatibility checks and its own validation evidence. Changes to a model, prompt, knowledge base, simulation solver or rule set should create a new skill version whose downstream dependencies can be identified before release. This turns modularity from a software convenience into a clinically governed property. Figure 1 summarizes the transition from isolated models to a governed clinical AI skill ecosystem.

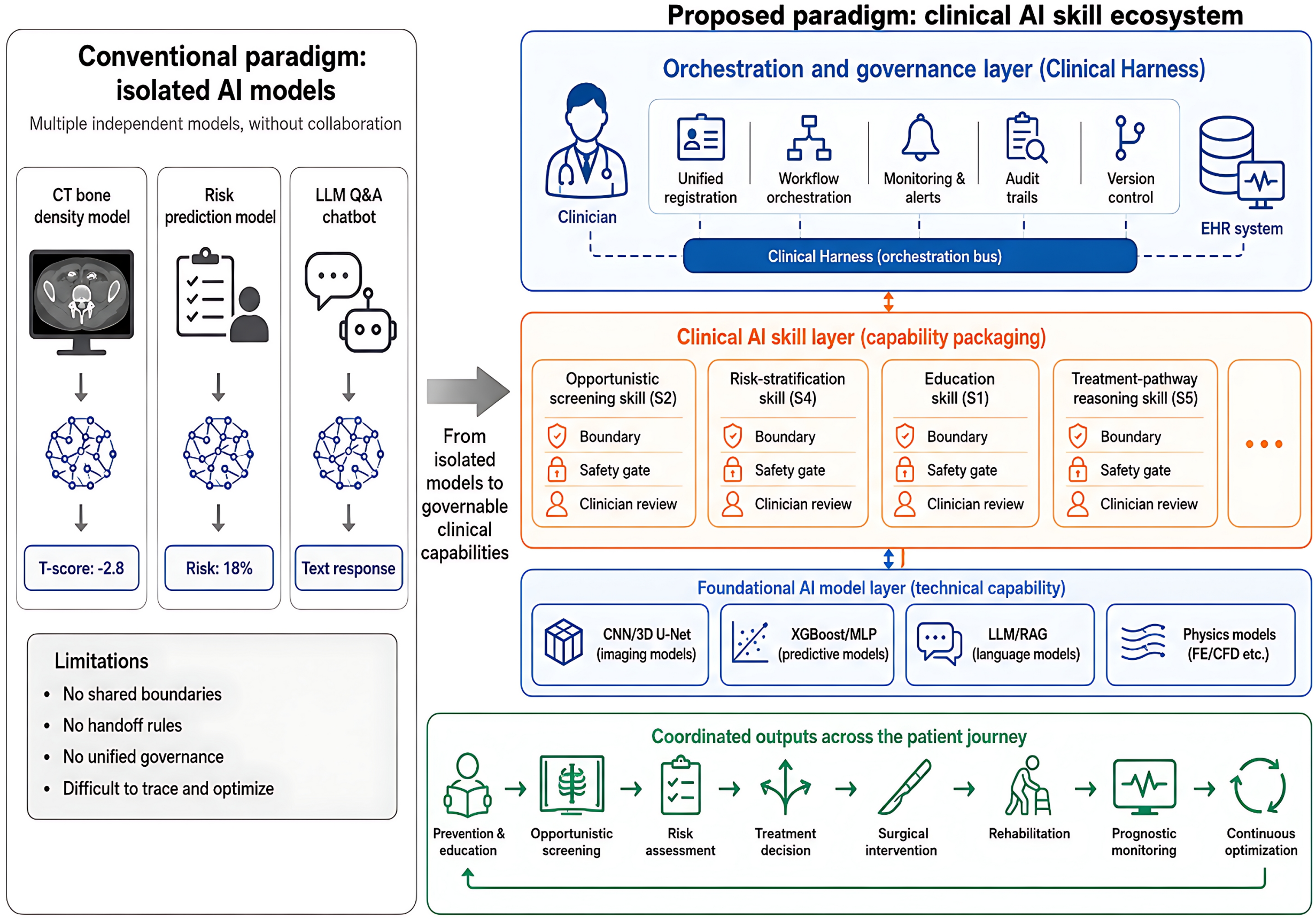

**Figure 1** | Conceptual evolution from isolated AI models to governable clinical AI skill ecosystems. In the conventional paradigm, independent prediction models are deployed without shared boundaries, handoff rules or governance. In the proposed paradigm, models are packaged as clinical AI skills and coordinated by the Clinical Harness across the patient journey.

# 3. Osteoporosis as an exemplar clinical AI skill ecosystem

Osteoporosis provides a demanding exemplar because it is common, chronic and underdiagnosed, and its management crosses prevention, screening, diagnosis, treatment selection, intervention, rehabilitation and monitoring. It therefore tests whether heterogeneous AI capabilities can be organized across a complete care pathway without obscuring their distinct evidence bases or responsibility boundaries.

The ecosystem contains nine clinical AI skills, each registered with explicit inputs, outputs, evidence status, boundaries, review requirements and monitoring rules. S1 provides guideline-grounded prevention and education through RAG and LLMs [6–8]. S2 supports opportunistic screening from routine CT, including bone-density estimation and vertebral-fracture detection [3,4,19]. S3 combines clinical evidence and imaging findings for diagnostic support in complex cases [20–23]. S4 integrates imaging, history, clinical indicators and treatment information into actionable fracture-risk strata [5,24,25].

S5 is a treatment-pathway reasoning skill that combines guideline logic with patient-specific evidence to propose medication, follow-up or referral, subject to mandatory clinician review [24,25]. S6 and S7 are physics-enhanced surgical skills. S6 combines vertebral segmentation, finite-element analysis and AI surrogate models to support cement-material and injection-volume planning [26–28]. S7 supports pedicle-access planning and cement-dispersion or leakage-risk assessment [28,29]. S8 supports rehabilitation and adherence management. S9 integrates bone density, pain, function, adherence and refracture events into a longitudinal outcome-risk trajectory [30].

The nine-skill ecosystem is an architectural specification, not a claim that nine validated modules already operate autonomously. Its purpose is to expose interfaces that isolated models leave implicit. S9 shows why those interfaces matter: refracture, bone-density decline, persistent pain, loss of function or poor adherence can reopen risk stratification, trigger pathway review and flag upstream skills for reassessment. Accountable transitions, rather than a catalogue of AI functions, organize the ecosystem. Supplementary Figure 1 maps the S1-S9 exemplar.

Comparable contracts are required in other longitudinal conditions. In heart failure, for example, imaging, risk estimation, medication titration and remote monitoring would use different skills but similar registration, handoff and review rules. The transferable element is the explicit definition of capability boundaries, dependencies and outcome feedback. Osteoporosis combines chronic management, opportunistic imaging and an interventional branch in a single exemplar.

## 4. Orchestration and hybrid reasoning: from prediction to action

Clinical value depends on orchestration as well as individual skill performance. Allowing an LLM to choose skill calls freely can create unstable paths and ambiguous accountability, particularly when clinical consequences accumulate across steps [31–35]. Deterministic clinical orchestration instead uses predefined pathways, conditional branches and human-review nodes to determine when each skill may be invoked.

The pathway is represented as a directed clinical graph and finite-state machine rather than an unconstrained natural-language plan. Each node has defined inputs, outputs, fallback conditions and review requirements. Structured artifact contracts carry measurements, provenance, units, confidence and validation status between skills. Natural language may explain a result, but it is not the sole control channel.

A representative workflow begins with routine CT. S2 estimates osteoporosis risk; low-risk cases may receive S1 education, whereas high-risk cases invoke S1, S3 and S4 in parallel. S5 combines their structured outputs with contraindications and patient preferences. A clinician then confirms, modifies or rejects the pathway. Conservative care proceeds to S8 and S9; surgical care may invoke S6 and S7 before S9 monitors outcomes and, when predefined triggers are met, reactivates S4 and S5. Supplementary Figure 2 shows this illustrative graph.

Here, deterministic orchestration means that permitted branches, exceptions and escalation routes are explicit and versioned. Conflicting outputs trigger reconciliation or review rather than silent averaging, while an unregistered input or unsupported transition stops the pathway. Hybrid reasoning remains reviewable, with clinical responsibility clearly assigned.

Concurrency also requires policy. Parallel skills may finish at different times, return discordant confidence estimates or depend on stale artifacts. The orchestration layer should specify which outputs are mandatory, how long they remain valid, whether one skill may supersede another and which disagreements require clinician arbitration. These are pathway semantics, not prompts, and should be testable before clinical deployment.

## 5. Physics-enhanced surgical AI and outcome closure

Physics-enhanced skills extend medical AI from diagnosis toward intervention. S6 and S7 combine finite-element and fluid-dynamic reasoning with AI-assisted segmentation and surrogate models [26–29]. In vertebral augmentation, anatomy, bone quality, cement properties, injection volume, dispersion and leakage risk interact. Physics-based simulation can represent constraints that a purely statistical predictor may miss, while surrogate models may reduce computation to workflow-compatible timescales.

Simulation evidence is not equivalent to clinical benefit. Segmentation errors, incorrect boundary conditions and unsuitable material assumptions can invalidate a result, and surrogate accuracy requires validation against both high-fidelity simulation and clinical outcomes. The Clinical Harness treats a physics-enhanced recommendation as a bounded planning artifact requiring review, rather than an autonomous surgical instruction.

S9 adds longitudinal outcome closure. Follow-up signals can update risk stratification, pathway selection and skill monitoring. Evaluation then moves beyond the correctness of one prediction to whether the governed workflow reduces fracture burden, preserves function and detects harm over time. Outcome closure supports both care and post-deployment governance.

## 6. Clinical Harness as runtime governance infrastructure

Runtime controls are needed to translate governance requirements into practice. Reporting and governance frameworks specify what should be documented, validated and monitored [15–18], but cannot themselves block an invalid input, stop an off-label invocation or require review before a high-risk action. The Clinical Harness turns those requirements into executable, auditable and interruptible controls.

The Clinical Harness occupies a distinct layer between AI capabilities and care execution. It can integrate with predictive models, electronic health records, workflow engines and MLOps platforms, while checking every clinically meaningful skill invocation against registered boundaries, pathway state, safety rules and human authority.

Agent research describes interleaved reasoning and action, tool use, reflection and iterative feedback [33–35]. Clinical use requires these mechanisms to remain outcome-anchored, human-governed and audit-preserving. Box 1 relates them to constrained care-execution, outcome-feedback and governance-improvement loops.

The architecture has four layers in the execution-oriented order shown in Figure 2. First, multilayer safety guardrails validate input completeness, format, range and provenance; enforce hard clinical constraints; monitor confidence; block requests outside registered scope; and require clinician confirmation for high-risk decisions. A guardrail is not a generic warning: it must have a defined trigger, response, owner and audit event. These controls address known risks from regulatory gaps, automation bias and generalist AI use [36–39]. Supplementary Figure 3 expands the guardrail sequence.

Second, the skill registry and boundary-management layer stores each skill card, including responsibility boundary, input-output contract, version, units, intended use, contraindications, validation status, failure modes and degradation strategy. Registration is a permissioning step, not a catalogue entry. It prevents unregistered or off-label execution and gives the guardrail layer the metadata needed to determine whether a requested action is allowed.

Third, the orchestration and state-management layer schedules skill invocation through versioned clinical graphs and finite-state transitions. Structured artifacts, rather than free-form conversational context alone, carry data and decisions across nodes. This layer records why a branch was taken, manages timeouts and unavailable skills, and routes conflicts or exceptions to predefined fallback and review paths.

Fourth, the audit and continuous-monitoring layer records user identity, input-data hashes, skill and rule versions, outputs, guardrail events, clinician actions, timestamps and outcome signals. Append-only logs support traceability and also drive runtime governance: repeated overrides, performance drift, subgroup bias or safety events can trigger alerts, restriction, suspension or revalidation [10,11,40]. Multicentre assessment can be linked to federated benchmarking without centralizing sensitive data [41].

The four layers operate as a single control system. Safety decisions depend on registry metadata; orchestration exposes the current state; audit evidence updates deployment status; and lifecycle governance spans design, validation, deployment, updating and retirement. The Clinical Harness cannot make an unvalidated model safe or eliminate distribution shift; rather, its testable purpose is to make procedural governance rules observable and enforceable during execution.

Governance policies should be represented as machine-checkable rules linked to human-readable rationale. A CT screening skill, for example, can be blocked when acquisition metadata fall outside its validated range; a treatment-pathway skill can require explicit contraindication fields; and a surgical-planning skill can be restricted when segmentation uncertainty exceeds a prespecified threshold. Each policy should identify its owner, required review, fallback action and supporting evidence, making the guardrail inspectable.

Box 1 | From loop engineering to human-governed clinical loops

Clinical Harness constrains iterative agent mechanisms through registered skills, predefined state transitions, structured artifacts and clinician authority.

| **General loop engineering** | **Clinical Harness interpretation** |
|---|---|
| Agent loop | Care-execution loop through registered skills |
| Verification loop | Safety loop through guardrails and review |
| Event-driven loop | Outcome-feedback loop triggered by clinical events |
| Self-improvement loop | Governance-improvement loop through audit and revalidation |

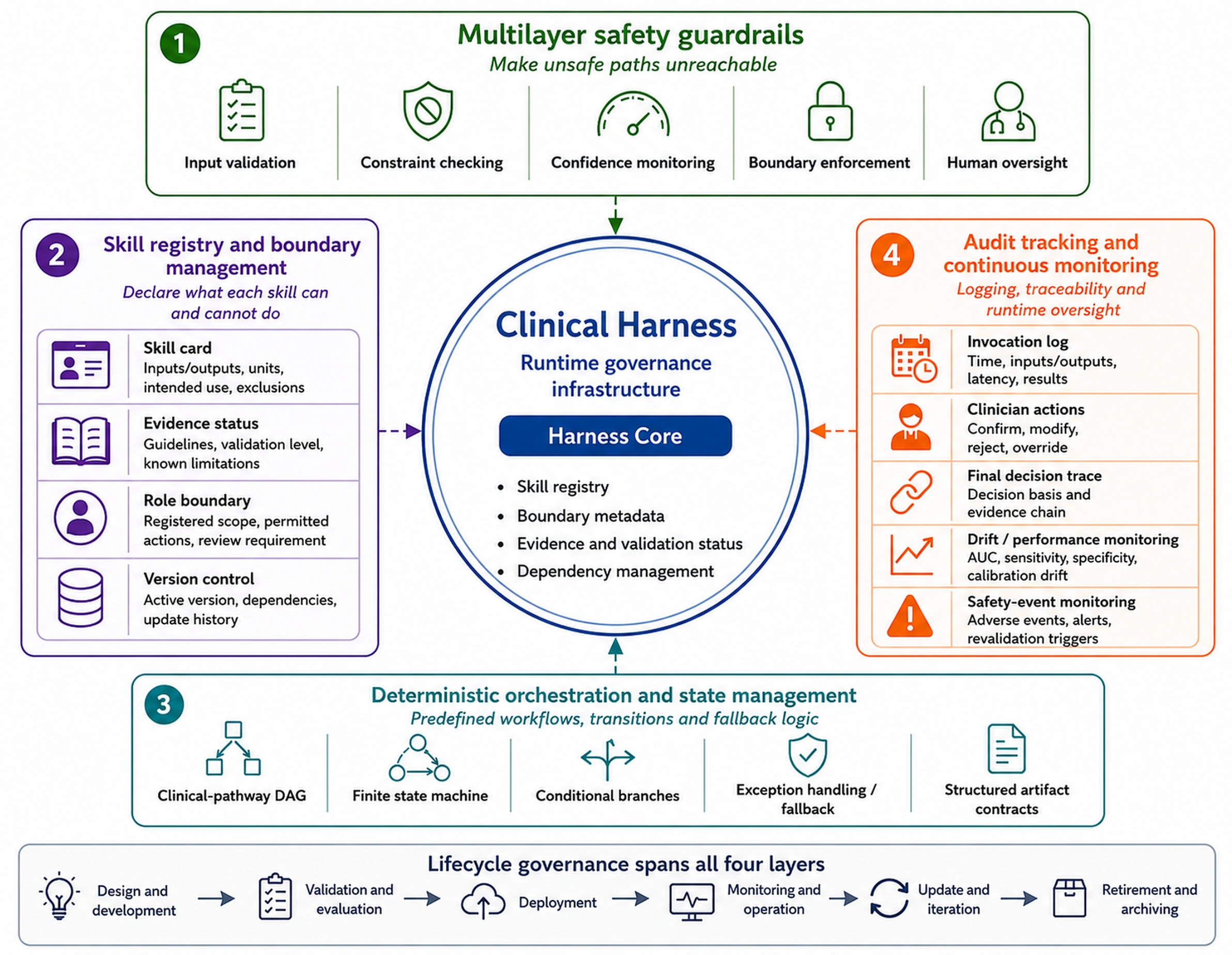

**Figure 2** | Four-layer runtime governance architecture of the Clinical Harness. The execution-oriented order is multilayer safety guardrails, skill registry and boundary management, deterministic orchestration with state management, and audit tracking with continuous monitoring. The Harness Core links skill registry, boundary metadata, evidence and validation status, and dependency management. Lifecycle governance spans design, validation, deployment, updating and retirement.

## 7. From skill ecosystems to governable medical agents

Medical agents require a constrained execution substrate rather than unrestricted tool use. Generalist and multimodal models may support broader reasoning [42,43], but do not inherently carry the intended-use boundaries, safety constraints or accountability required for clinical action. The Clinical Harness provides one possible substrate for governable medical agents.

Within this arrangement, an agent plans above registered, validated and monitored clinical AI skills. It may propose a next step, but the Clinical Harness determines whether the step is permitted, what evidence is required, which artifact contract applies and when a clinician must intervene. State transitions and audit logs make the sequence traceable.

A governable medical agent is not equivalent to the Clinical Harness. The agent provides reasoning and action; the harness provides runtime constraint and accountability. Figure 3 distinguishes these roles and treats autonomy as a bounded, revocable permission rather than a permanent property of a model.

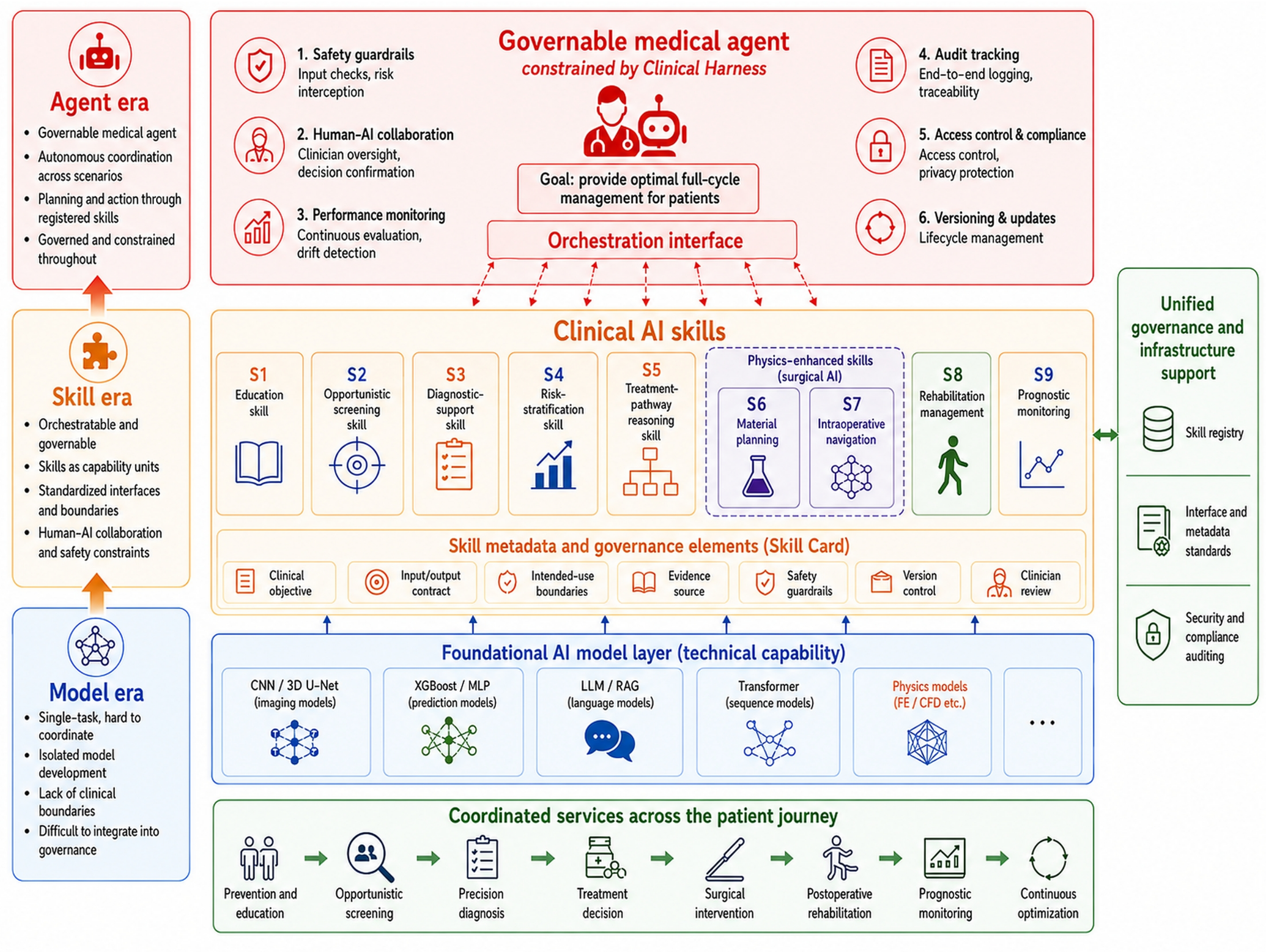

**Figure 3** | Evolution from isolated models to governable medical agents. A governable medical agent is not equivalent to the Clinical Harness; it invokes standardized clinical AI skills whose behaviour is constrained through registered boundaries, safety gates, state transitions, audit trails and runtime governance.

## 8. Implementation and validation roadmap

Implementation should proceed incrementally rather than begin with an autonomous clinical system. Four evidence-gated phases move from component validation to monitored clinical use while clinicians retain control.

First, each skill should be validated before registration. Knowledge-driven skills require tests of guideline fidelity, evidence attribution, unsafe recommendation rate and subgroup appropriateness. Data-driven skills require discrimination, calibration, robustness, missing-data behaviour and decision-threshold analysis. Physics-enhanced skills require segmentation reliability, surrogate accuracy, planning reproducibility and explicit failure testing. The skill card should record exclusions, evidence, fallback rules and required oversight [15–18].

Second, the harness should enter silent deployment. It runs without changing care while recording which skills would have been invoked, whether artifact contracts were complete, which guardrails fired, where escalation occurred and how often the proposed pathway disagreed with clinicians. This tests workflow fit, latency, data availability, audit completeness and safety-gate behaviour before recommendations are exposed.

Third, prospective workflow evaluation should assess clinical utility with clinicians retaining authority, using a pragmatic trial, stepped-wedge design or prospective implementation study as appropriate. Endpoints should include care-process measures, override and escalation patterns, safety, patient-reported outcomes and workload, not accuracy alone. Fourth, post-deployment monitoring should track drift, subgroup performance, repeated overrides, safety events, version changes and guideline updates, with predefined triggers for recalibration, suspension, revalidation or retirement [10,11,41]. Supplementary Table 2 provides an illustrative roadmap.

Evaluation must treat the Clinical Harness as a sociotechnical intervention. A technically correct guardrail can still fail if it creates alert burden, an audit trail can be complete yet unusable, and a high-performing skill can worsen inequity when embedded in an inaccessible pathway. The central question is whether governed capability organization improves decisions and outcomes without shifting hidden work or risk to clinicians and patients.

Movement between phases should be evidence-gated. Registration does not authorize live use, and successful silent deployment does not by itself justify prospective activation. Each transition requires predefined acceptance criteria, named clinical and technical owners, rollback procedures and documentation of unresolved risks. This establishes a traceable path from model evidence to pathway and outcome evidence for regulatory and institutional review.

## 9. Conclusion

Translating validated models into safe clinical workflows requires explicit organization of capability, evidence and responsibility. Osteoporosis is the exemplar, but the runtime architecture is disease-agnostic. Clinical AI skills make intent, evidence and boundaries explicit; deterministic orchestration connects them across time; and outcome closure links care to monitoring.

Within this architecture, the Clinical Harness expresses governance as executable constraints, logs, monitoring and intervention points. General agent loops become human-governed clinical loops in which execution, feedback and system improvement remain interruptible and auditable.

The same organizing principles apply to longitudinal conditions such as heart failure, diabetes and chronic obstructive pulmonary disease. They define a path from isolated models to clinical AI skills, from skills to governed workflows and from governed workflows to medical agents operating within explicit constraints.

## Acknowledgements

This work was supported by the Jiangsu Provincial Frontier Technology Research and Development Program (Department of Science and Technology of Jiangsu Province; grant no. BF2024044).

## Author contributions

T.X.: conceptualization and writing - original draft. L.B.: conceptualization and writing - review and editing. Z.H.: writing - review and editing. T.S.: writing - review and editing. Y.W.: conceptualization, supervision and writing - review and editing. All authors reviewed and approved the final manuscript.

## Competing interests

The authors declare no competing interests.

## Data availability

Data availability is not applicable as no datasets were generated or analysed in this article.

## Code availability

Code availability is not applicable as no software code was generated for this article.

# Supplementary Information

**Supplementary Table 1 |** Clinical AI Skill Cards for S1-S9.

Registration requires each skill card to specify the clinical objective, intended-use population, inputs, outputs, operating boundary, exclusions or contraindications, validation evidence, safety gates and fallback, human oversight, and audit and monitoring. All fields require local validation against clinical pathways, data and regulatory requirements before deployment.

| Skill | Card field | Specification |
|---|---|---|
| **S1 Prevention and education skill** | Clinical objective | Deliver individualized bone-health education, lifestyle guidance and long-term self-management support for the public and high-risk groups. |
| | Intended-use population | Public users, people at elevated osteoporosis risk, and patients requiring bone-health education or adherence support. |
| | Inputs | Patient age, sex, fracture history, osteoporosis risk factors, lifestyle profile, medication history, preferences, health-literacy level and retrieved guideline/consensus evidence. |
| | Outputs | Evidence-grounded education summary, lifestyle recommendations, fall-prevention advice, adherence prompts and suggested topics for clinician discussion. |
| | Operating boundary | Education and self-management support within validated guideline-grounded content and local clinical pathways. |
| | Exclusions / contraindications | Does not diagnose osteoporosis, prescribe medication, change treatment plans or manage emergency symptoms. Escalates when red flags, severe pain, new neurological deficit or suspected acute fracture are reported. |
| | Validation evidence | Guideline fidelity review, expert assessment of education quality, hallucination/unsafe-advice testing, readability testing and subgroup appropriateness assessment. |
| | Safety gates and fallback | Retrieval-grounding requirement, source attribution, unsafe recommendation filter, medication/supplement caution check and fallback to clinician or health educator when confidence or evidence quality is low. |
| | Human oversight | Human-on-the-loop for routine education; clinician review required for medication-related or high-risk advice. |
| | Audit and monitoring | Records source documents, prompt/context, generated advice, safety flags, clinician or patient feedback, revision history and adverse feedback patterns. |
| **S2 Opportunistic screening skill** | Clinical objective | Estimate bone mineral density and osteoporosis risk from routine CT imaging to support case finding without additional radiation exposure. |
| | Intended-use population | Patients undergoing routine CT scans that include validated vertebral regions and acquisition protocols. |
| | Inputs | Routine CT DICOM data, scanner and protocol metadata, vertebral region of interest, image quality indicators and relevant demographic variables. |
| | Outputs | Estimated BMD/T-score or risk category, confidence/uncertainty score, image-quality flag and recommendation for DXA or clinical assessment when appropriate. |
| | Operating boundary | Screening aid only, applicable to validated CT protocols, anatomical coverage, scanner settings and slice-thickness ranges. |
| | Exclusions / contraindications | Excludes severe artifacts, hardware obscuring the vertebrae, failed segmentation or out-of-distribution protocols. |
| | Validation evidence | External validation against DXA or accepted reference standards, calibration assessment, scanner/vendor robustness, subgroup analysis and normative-value benchmarking. |
| | Safety gates and fallback | Input-quality check, segmentation quality control, uncertainty threshold, protocol compatibility check and automatic withholding when image quality or model confidence is insufficient. |

| Skill | Card field | Specification |
|---|---|---|
| | Human oversight | Radiologist or responsible clinician confirms clinical relevance; AI output is a screening aid rather than a diagnostic endpoint. |
| | Audit and monitoring | Records image hash, scanner/protocol metadata, model version, BMD/risk output, uncertainty, quality flags, referral recommendation and downstream DXA or diagnostic confirmation. |
| **S3 Diagnostic-support skill** | Clinical objective | Support clinicians in diagnostic reasoning for osteoporosis, vertebral fracture and complex comorbidity scenarios. |
| | Intended-use population | Patients with suspected or confirmed osteoporosis, vertebral fracture, atypical presentation or complex comorbidity requiring diagnostic reasoning support. |
| | Inputs | Symptoms, imaging findings, laboratory results, medication history, comorbidities, prior fractures, clinician question and retrieved guideline/literature/case evidence. |
| | Outputs | Evidence-linked diagnostic considerations, suggested work-up, differential diagnosis support, uncertainty notes and cited sources for clinician review. |
| | Operating boundary | Diagnostic reasoning support that organizes evidence and highlights uncertainty; final diagnostic responsibility remains with clinicians. |
| | Exclusions / contraindications | Does not make final diagnosis or replace specialist judgement. Escalates rare, inconsistent, urgent or low-evidence cases. |
| | Validation evidence | Clinician evaluation on representative cases, citation accuracy, conflict-detection testing, rare-case review and guideline-concordance assessment. |
| | Safety gates and fallback | Mandatory source citation, evidence-level tagging, contradiction detection, hallucination screening and fallback to clinician when evidence is insufficient or conflicting. |
| | Human oversight | Clinician-in-the-loop; all diagnostic outputs require clinician interpretation and confirmation. |
| | Audit and monitoring | Records clinical query, retrieved evidence, output rationale, evidence conflicts, uncertainty tags, clinician acceptance/modification/rejection and subsequent diagnosis. |
| **S4 Risk-stratification skill** | Clinical objective | Estimate future fracture risk and disease-progression probability and assign patients to actionable management strata. |
| | Intended-use population | Patients with sufficient imaging, clinical and historical data who require fracture-risk or disease-progression stratification. |
| | Inputs | Imaging-derived features, BMD or CT-derived bone metrics, prior fracture history, age, sex, comorbidities, medications, laboratory data, fall risk, treatment history and follow-up data. |
| | Outputs | Risk score, calibrated probability, risk stratum, explanatory feature summary and recommendation for pathway review when thresholds are met. |
| | Operating boundary | Risk estimation within validated populations, data-completeness thresholds and prespecified decision thresholds. |
| | Exclusions / contraindications | Flags missing or out-of-distribution inputs and does not directly prescribe treatment. |
| | Validation evidence | Discrimination, calibration, decision-curve analysis, subgroup robustness, temporal validation, external validation and threshold-sensitivity testing. |
| | Safety gates and fallback | Missing-data warning, calibration/uncertainty check, out-of-distribution detection and escalation when risk estimates are unstable or clinically discordant. |
| | Human oversight | Clinician interprets risk in context of patient preferences, competing risks and treatment contraindications. |
| | Audit and monitoring | Records input variables, missingness, risk estimate, calibration status, thresholds used, model version, clinician override and observed fracture outcomes. |

| Skill | Card field | Specification |
| --- | --- | --- |
| **S5 Treatment-pathway reasoning skill** | Clinical objective | Translate diagnostic findings, risk status, contraindications and patient preferences into individualized treatment-pathway recommendations. |
| | Intended-use population | Patients who have completed diagnostic and risk assessment and require pharmacological, non-pharmacological, follow-up or referral pathway review. |
| | Inputs | S3 diagnostic support output, S4 risk stratum, medication history, renal function and relevant laboratory results, contraindications, prior adverse events, patient preference and guideline/case evidence. |
| | Outputs | Suggested medication class or non-pharmacological pathway, follow-up interval, referral recommendation and rationale with evidence and contraindication checks. |
| | Operating boundary | Recommendation support only, operating within guideline logic, local pathways and clinician-defined constraints. |
| | Exclusions / contraindications | Does not autonomously place orders, prescribe medication or perform dose adjustment. High-risk or conflicting scenarios require clinician decision. |
| | Validation evidence | Guideline concordance, expert panel review, contraindication-detection testing, case-based pathway evaluation and monitoring of clinician acceptance and overrides. |
| | Safety gates and fallback | Hard contraindication and allergy checks, renal-function constraints, drug-safety checks, evidence-level tagging, conflict detection and mandatory review for consequential treatment decisions. |
| | Human oversight | Mandatory clinician-in-the-loop before treatment initiation, medication changes or surgical referral. |
| | Audit and monitoring | Records recommendation, evidence sources, contraindication checks, clinician action, override reason, final treatment decision and downstream outcome/adverse events. |
| **S6 Material-planning skill** | Clinical objective | Support personalized vertebral augmentation planning by estimating biomechanical status and recommending cement-related parameters. |
| | Intended-use population | Patients being considered for vertebral augmentation with adequate preoperative CT and validated vertebral anatomy/procedure context. |
| | Inputs | Preoperative CT, vertebral segmentation, fracture morphology, bone-quality metrics, planned procedure context and finite-element/surrogate-model parameters. |
| | Outputs | Estimated biomechanical status, suggested cement elastic modulus, injection-volume range and augmentation strategy for surgeon review. |
| | Operating boundary | Adjunct to preoperative planning only, under surgeon review and within validated simulation/model assumptions. |
| | Exclusions / contraindications | Excludes poor CT quality, failed segmentation, atypical anatomy outside the validated domain or cases requiring alternative surgical judgement. |
| | Validation evidence | Segmentation reliability, finite-element model verification, surrogate-model accuracy, reproducibility testing and retrospective/prospective comparison with surgical outcomes. |
| | Safety gates and fallback | Image-quality and segmentation-quality checks, parameter-range constraints, uncertainty threshold, anatomy/outlier detection and fallback to conventional planning when validation conditions are not met. |
| | Human oversight | Mandatory surgeon review and confirmation; AI output is not executable surgical instruction. |
| | Audit and monitoring | Records CT hash, segmentation version, simulation/model version, planning parameters, recommendation, uncertainty, surgeon edits and postoperative complications or cement-related outcomes. |
| **S7 Intraoperative navigation skill** | Clinical objective | Assist trajectory planning and cement-dispersion/leakage risk assessment during vertebral augmentation. |

| Skill | Card field | Specification |
|---|---|---|
| | Intended-use population | Patients undergoing vertebral augmentation in workflows with validated intraoperative imaging, registration and navigation data. |
| | Inputs | Preoperative plan, intraoperative imaging or navigation data, vertebral anatomy, pedicle geometry, planned access trajectory, cement properties and injection context when available. |
| | Outputs | Candidate pedicle access trajectory, anatomical risk alerts, predicted cement-dispersion, leakage-risk estimate and intraoperative warning for clinician review. |
| | Operating boundary | Decision support within validated navigation and imaging conditions; continuous clinician control is required. |
| | Exclusions / contraindications | Does not autonomously control instruments or inject cement. Excludes inadequate imaging, registration failure, unstable anatomy or conditions outside validated workflow. |
| | Validation evidence | Trajectory reproducibility, registration accuracy, leakage prediction performance, prospective usability testing and correlation with cement leakage or procedural complications. |
| | Safety gates and fallback | Registration and image-quality checks, trajectory boundary validation, leakage-risk alert thresholds, fail-safe interruption and immediate escalation when the predicted risk exceeds predefined limits. |
| | Human oversight | Continuous clinician control; surgeon confirms, modifies or ignores all navigation suggestions. |
| | Audit and monitoring | Records imaging/registration data, planned and modified trajectories, risk maps, alerts, timing, clinician actions, cement leakage events and peri-procedural complications. |
| **S8 Rehabilitation-management skill** | Clinical objective | Generate individualized rehabilitation, lifestyle and adherence-management support after diagnosis, treatment or surgery. |
| | Intended-use population | Patients after diagnosis, pharmacological treatment or surgery who require rehabilitation planning, lifestyle support or adherence management. |
| | Inputs | Treatment-pathway, procedure type, pain score, functional status, fall risk, medication adherence, comorbidities, rehabilitation contraindications, patient goals and guideline evidence. |
| | Outputs | Rehabilitation plan, exercise progression, lifestyle advice, adherence reminders, red-flag education and follow-up prompts. |
| | Operating boundary | Rehabilitation and adherence support within validated guidelines, patient status and clinician-defined restrictions. |
| | Exclusions / contraindications | Does not manage emergencies, replace physiotherapist/clinician assessment or recommend exercises in the presence of red-flag symptoms without review. |
| | Validation evidence | Guideline fidelity, physiotherapist/clinician review, safety testing for contraindicated exercises, adherence impact and patient usability/readability assessment. |
| | Safety gates and fallback | Red-flag symptom detection, contraindication checks, pain/function threshold monitoring, evidence attribution and escalation when symptoms worsen or the plan conflicts with clinical restrictions. |
| | Human oversight | Clinician or rehabilitation specialist reviews high-risk rehabilitation plans; routine adherence support can be monitored human-on-the-loop. |
| | Audit and monitoring | Records plan version, exercise recommendations, patient-reported symptoms, adherence, red-flag alerts, clinician modifications and functional outcomes. |
| **S9 Prognostic-monitoring skill** | Clinical objective | Continuously monitor long-term outcomes and trigger reassessment when disease control, function or adherence deteriorates. |
| | Intended-use population | Patients in follow-up after screening, diagnosis, treatment, surgery or rehabilitation with sufficient longitudinal outcome data. |

| Skill | Card field | Specification |
| --- | --- | --- |
| | Inputs | Follow-up BMD or CT-derived metrics, pain score, function, falls, refracture events, medication adherence, adverse events, imaging, laboratory data and patient-reported outcomes. |
| | Outputs | Longitudinal risk trajectory, deterioration alert, follow-up recommendation, re-stratification trigger and performance feedback signal for upstream skills. |
| | Operating boundary | Monitoring and triage support based on longitudinal data and predefined alert thresholds. |
| | Exclusions / contraindications | Requires sufficient longitudinal data and does not replace urgent clinical evaluation when acute events are reported. |
| | Validation evidence | Temporal calibration, prediction of refracture or functional decline, alert sensitivity/specificity, subgroup performance and evaluation of alarm burden. |
| | Safety gates and fallback | Data-completeness check, trend-validation rules, alert-priority thresholds, alarm-fatigue monitoring and escalation for refracture, rapid BMD decline, persistent pain or functional deterioration. |
| | Human oversight | Clinician reviews clinically significant alerts and determines whether to reopen S4 risk stratification, S5 pathway reasoning or surgical/rehabilitation review. |
| | Audit and monitoring | Records longitudinal inputs, alert thresholds, generated alerts, clinician response, downstream outcomes, drift signals, override patterns and revalidation triggers. |

**Supplementary Table 2 |** Implementation and validation roadmap for the Clinical Harness, using osteoporosis care as an exemplar.

| Phase | Objective | Key validation focus | Example endpoints |
|---|---|---|---|
| **1. Skill-level validation** | Validate each S1-S9 skill before registration. | Input quality, accuracy, calibration, subgroup robustness, safety, failure modes and fallback rules. | Guideline fidelity; calibration; segmentation or surrogate-model accuracy. |
| **2. Silent deployment** | Run Clinical Harness in the background without affecting care. | Workflow fit, artifact completeness, guardrail activation, latency, audit-log completeness and clinician-disagreement patterns. | Skill-call rate; safety-gate trigger rate; escalation frequency. |
| **3. Prospective workflow evaluation** | Test clinical utility with clinicians in control. | Care-process improvement, safety, usability, clinician trust and patient-centred benefit. | Screening yield; treatment initiation; clinician override; patient-reported outcomes. |
| **4. Post-deployment monitoring** | Maintain performance and governance after deployment. | Drift, bias, safety events, version changes, guideline updates, revalidation and retirement triggers. | Calibration drift; subgroup gaps; adverse events; revalidation events. |

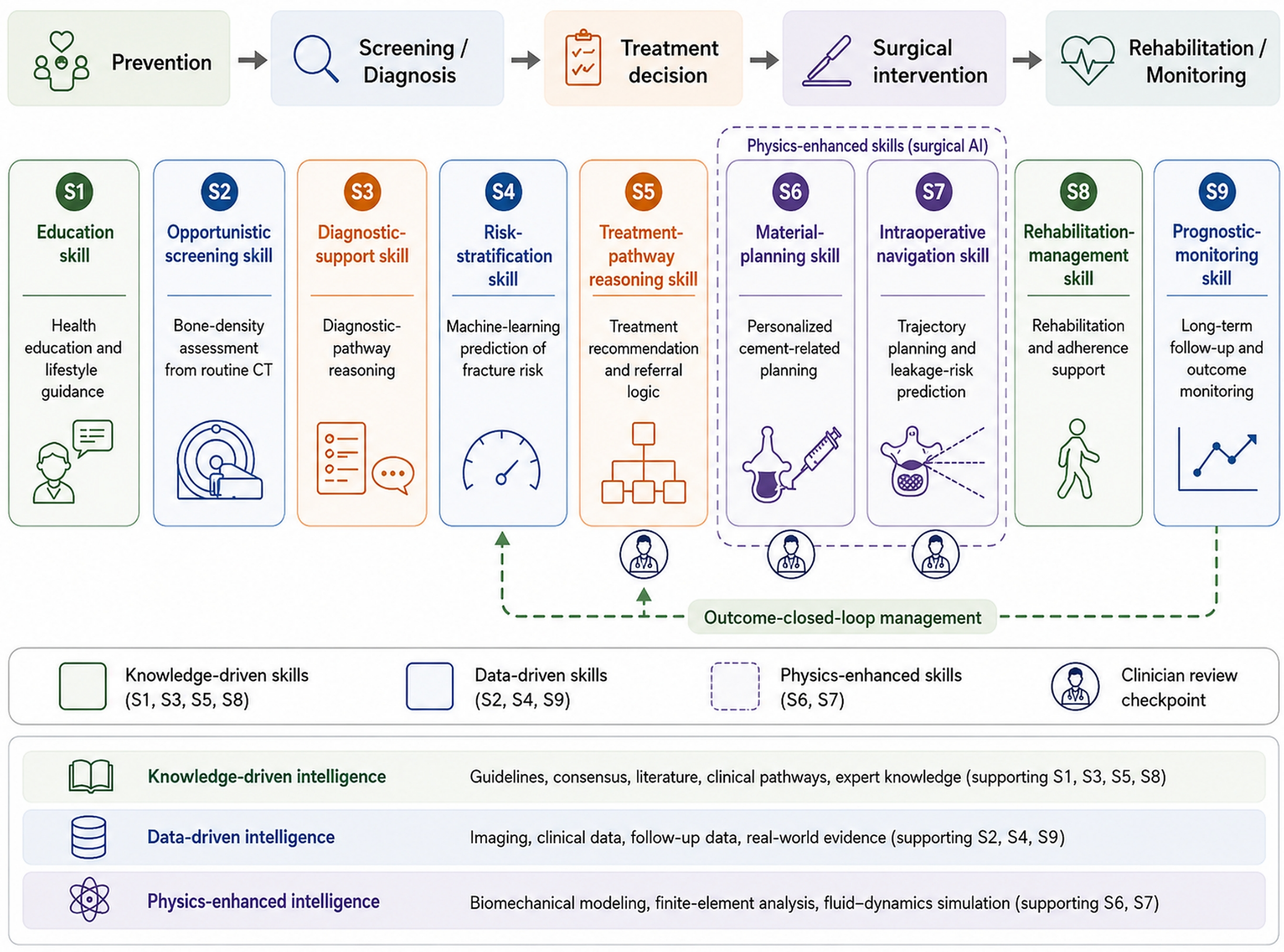

**Supplementary Figure 1 |** Exemplar clinical AI skills for osteoporosis lifecycle management. S1-S9 map to prevention, opportunistic screening, diagnostic support, risk stratification, treatment-pathway reasoning, surgical planning, intraoperative navigation, rehabilitation and longitudinal monitoring.

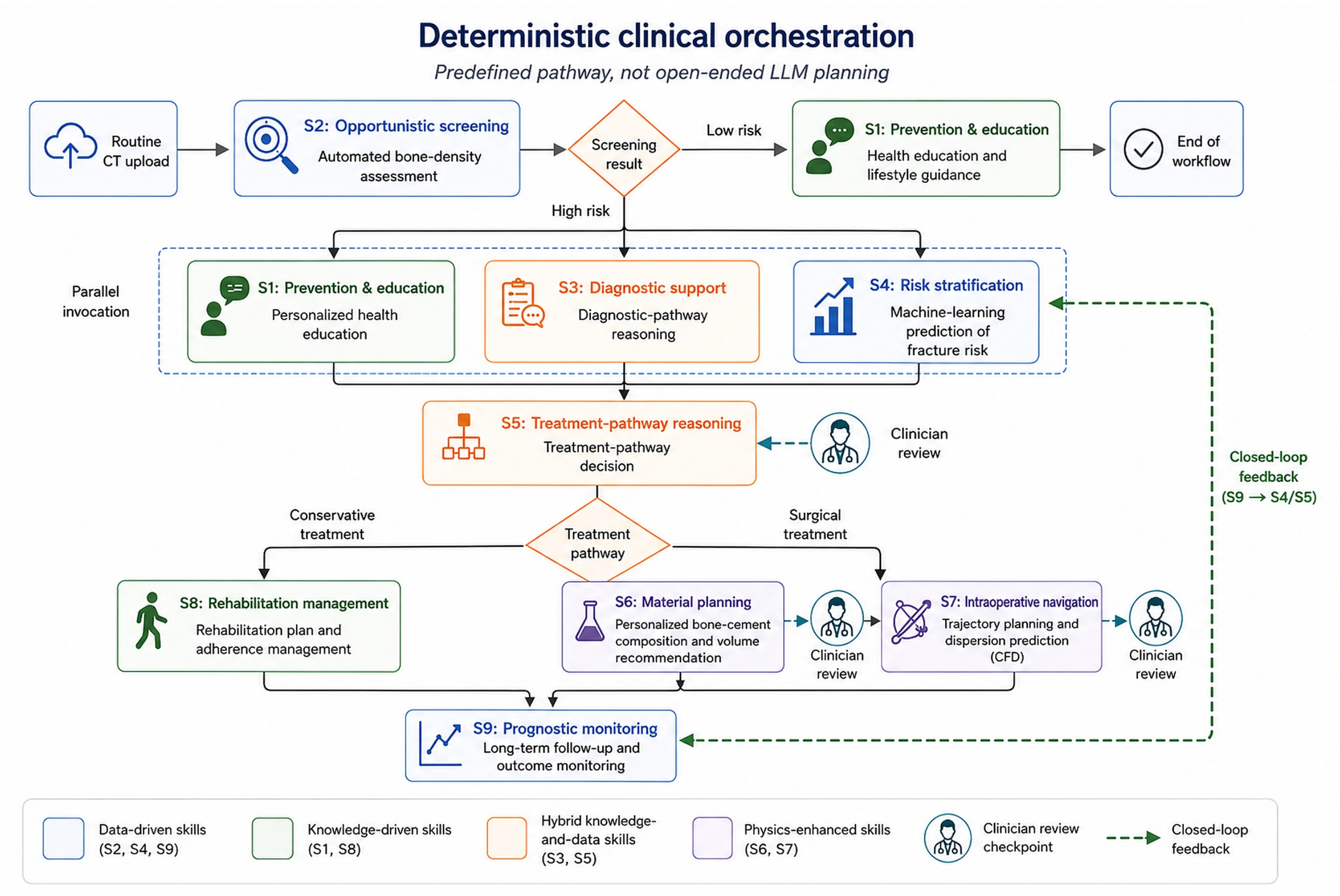

**Supplementary Figure 2 |** Illustrative deterministic orchestration graph. Routine CT initiates S2 screening, predefined branches invoke registered skills, clinician-review nodes constrain consequential decisions, and S9 outcome monitoring can reactivate S4 and S5.

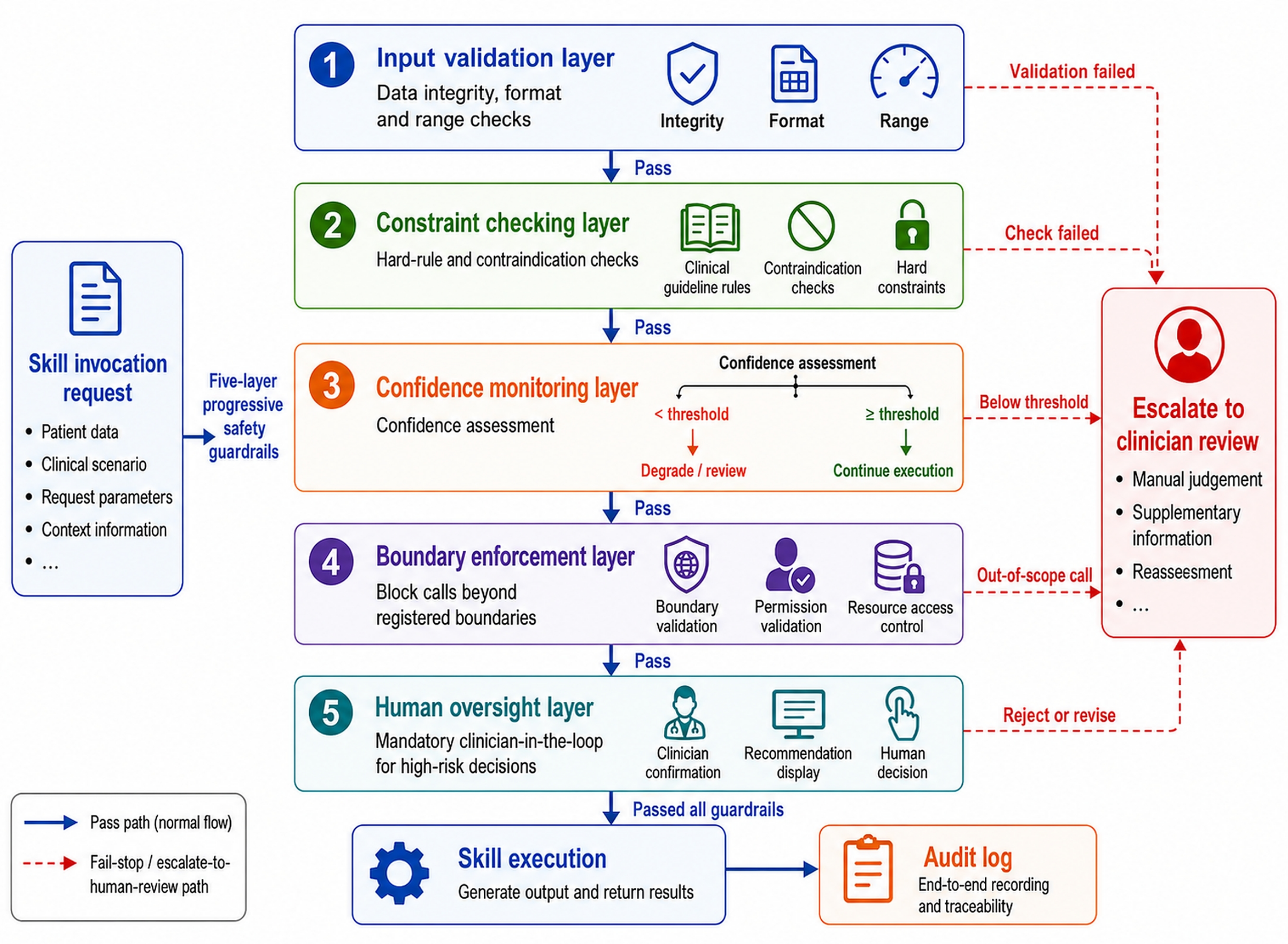

**Supplementary Figure 3 |** Multilayer safety guardrails. Invocation requests pass through input validation, hard-rule constraint checks, confidence monitoring, boundary enforcement and human oversight. Failure or uncertainty interrupts execution, routes review when appropriate and records an audit event.